# Deep Learning:
# A Critical Appraisal

## Gary Marcus[1]
## New York University

## Abstract

Although deep learning has historical roots going back decades, neither the term "deep learning" nor the approach was popular just over five years ago, when the field was reignited by papers such as Krizhevsky, Sutskever and Hinton's now classic 2012 (Krizhevsky, Sutskever, & Hinton, 2012)deep net model of Imagenet.

What has the field discovered in the five subsequent years? Against a background of considerable progress in areas such as speech recognition, image recognition, and game playing, and considerable enthusiasm in the popular press, I present ten concerns for deep learning, and suggest that deep learning must be supplemented by other techniques if we are to reach artificial general intelligence.

---

1 Departments of Psychology and Neural Science, New York University, gary.marcus at <u>nyu.edu</u>. I thank Christina Chen, François Chollet, Ernie Davis, Zack Lipton, Stefano Pacifico, Suchi Saria, and Athena Vouloumanos for sharp-eyed comments, all generously supplied on short notice during the holidays at the close of 2017.

> *For most problems where deep learning has enabled transformationally better solutions (vision, speech), we've entered diminishing returns territory in 2016-2017.*
>
> François Chollet, Google, author of Keras neural network library
> December 18, 2017

> *'Science progresses one funeral at a time.' The future depends on some graduate student who is deeply suspicious of everything I have said.*
>
> Geoff Hinton, grandfather of deep learning
> September 15, 2017

# 1. Is deep learning approaching a wall?

Although deep learning has historical roots going back decades(Schmidhuber, 2015), it attracted relatively little notice until just over five years ago. Virtually everything changed in 2012, with the publication of a series of highly influential papers such as Krizhevsky, Sutskever and Hinton's 2012 <u>ImageNet Classification with Deep Convolutional Neural Networks</u> (Krizhevsky, Sutskever, & Hinton, 2012), which achieved state-of-the-art results on the object recognition challenge known as ImageNet (Deng et al., ). Other labs were already working on similar work (Cireşan, Meier, Masci, & Schmidhuber, 2012). Before the year was out, deep learning made the front page of *The New York Times*[2], and it rapidly became the best known technique in artificial intelligence, by a wide margin. If the general idea of training neural networks with multiple layers was not new, it was, in part because of increases in computational power and data, the first time that deep learning truly became practical.

Deep learning has since yielded numerous state of the art results, in domains such as speech recognition, image recognition , and language translation and plays a role in a wide swath of current AI applications. Corporations have invested billions of dollars fighting for deep learning talent. One prominent deep learning advocate, Andrew Ng, has gone so far to suggest that "If a typical person can do a mental task with less than one second of thought, we can probably automate it using AI either now or in the near

---

[2] http://www.nytimes.com/2012/11/24/science/scientists-see-advances-in-deep-learning-a-part-of-artificial-intelligence.html

Page 2 of 27

future." (A, 2016). A recent New York Times Sunday Magazine article[3], largely about deep learning, implied that the technique is "poised to reinvent computing itself."

Yet deep learning may well be approaching a wall, much as I anticipated earlier, at beginning of the resurgence (Marcus, 2012), and as leading figures like Hinton (Sabour, Frosst, & Hinton, 2017) and Chollet (2017) have begun to imply in recent months.

What exactly is deep learning, and what has its shown about the nature of intelligence? What can we expect it to do, and where might we expect it to break down? How close or far are we from "artificial general intelligence", and a point at which machines show a human-like flexibility in solving unfamiliar problems? The purpose of this paper is both to temper some irrational exuberance and also to consider what we as a field might need to move forward.

This paper is written simultaneously for researchers in the field, and for a growing set of AI consumers with less technical background who may wish to understand where the field is headed. As such I will begin with a very brief, nontechnical introduction[4] aimed at elucidating what deep learning systems do well and why (Section 2), before turning to an assessment of deep learning's weaknesses (Section 3) and some fears that arise from misunderstandings about deep learning's capabilities (Section 4), and closing with perspective on going forward (Section 5).

Deep learning is not likely to disappear, nor should it. But five years into the field's resurgence seems like a good moment for a critical reflection, on what deep learning has and has not been able to achieve.

## 2. What deep learning is, and what it does well

Deep learning, as it is primarily used, is essentially a statistical technique for classifying patterns, based on sample data, using neural networks with multiple layers.[5]

---

[3] https://www.nytimes.com/2016/12/14/magazine/the-great-ai-awakening.html

[4] For more technical introduction, there are many excellent recent tutorials on deep learning including (Chollet, 2017) and (Goodfellow, Bengio, & Courville, 2016), as well as insightful blogs and online resources from Zachary Lipton, Chris Olah, and many others.

[5] Other applications of deep learning beyond classification are possible, too, though currently less popular, and outside of the scope of the current article. These include using deep learning as an alternative to regression, as a component in generative models that create (e.g.,) synthetic images, as a tool for compressing images, as a tool for learning probability distributions, and (relatedly) as an important technique for approximation known as variational inference.

Page 3 of 27

Neural networks in the deep learning literature typically consist of a set of input units that stand for things like pixels or words, multiple hidden layers (the more such layers, the deeper a network is said to be) containing hidden units (also known as nodes or neurons), and a set output units, with connections running between those nodes. In a typical application such a network might be trained on a large sets of handwritten digits (these are the inputs, represented as images) and labels (these are the outputs) that identify the categories to which those inputs belong (this image is a 2, that one is a 3, and so forth).

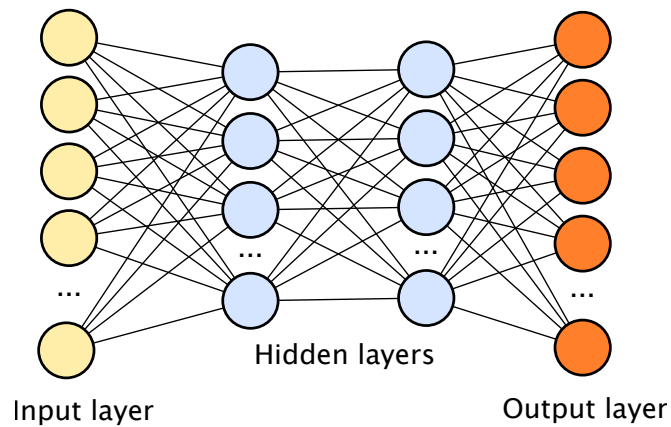

Over time, an algorithm called *back-propagation* allows a process called gradient descent to adjust the connections between units using a process, such that any given input tends to produce the corresponding output.

Collectively, one can think of the relation between inputs and outputs that a neural network learns as a *mapping*. Neural networks, particularly those with multiple hidden layers (hence the term *deep*) are remarkably good at learning input-output mappings,

Such systems are commonly described as neural networks because the input nodes, hidden nodes, and output nodes can be thought of as loosely analogous to biological neurons, albeit greatly simplified, and the connections between nodes can be thought of as in some way reflecting connections between neurons. A longstanding question, outside the scope of the current paper, concerns the degree to which artificial neural networks are biologically plausible.

Most deep learning networks make heavy use of a technique called *convolution* (LeCun, 1989), which constrains the neural connections in the network such that they innately capture a property known as translational invariance. This is essentially the idea that an object can slide around an image while maintaining its identity; a circle in the top left can be presumed, even absent direct experience) to be the same as a circle in the bottom right.

Deep learning is also known for its ability to self-generate intermediate representations, such as internal units that may respond to things like horizontal lines, or more complex elements of pictorial structure.

In principle, given infinite data, deep learning systems are powerful enough to represent any finite deterministic "mapping" between any given set of inputs and a set of corresponding outputs, though in practice whether they can learn such a mapping depends on many factors. One common concern is getting caught in *local minima*, in which a systems gets stuck on a suboptimal solution, with no better solution nearby in the space of solutions being searched. (Experts use a variety of techniques to avoid such problems, to reasonably good effect). In practice, results with large data sets are often quite good, on a wide range of potential mappings.

In speech recognition, for example, a neural network learns a mapping between a set of speech sounds, and set of labels (such as words or phonemes). In object recognition, a neural network learns a mapping between a set of images and a set of labels (such that, for example, pictures of cars are labeled as cars). In DeepMind's Atari game system (Mnih et al., 2015), neural networks learned mappings between pixels and joystick positions.

Deep learning systems are most often used as classification system in the sense that the mission of a typical network is to decide which of a set of categories (defined by the output units on the neural network) a given input belongs to. With enough imagination, the power of classification is immense; outputs can represent words, places on a Go board, or virtually anything else.

In a world with infinite data, and infinite computational resources, there might be little need for any other technique.

# 3. Limits on the scope of deep learning

Deep learning's limitations begin with the contrapositive: we live in a world in which data are never infinite. Instead, systems that rely on deep learning frequently have to generalize beyond the specific data that they have seen, whether to a new pronunciation of a word or to an image that differs from one that the system has seen before, and where data are less than infinite, the ability of formal proofs to guarantee high-quality performance is more limited.

As discussed later in this article, generalization can be thought of as coming in two flavors, *interpolation* between known examples, and *extrapolation,* which requires going beyond a space of known training examples (Marcus, 1998a).

For neural networks to generalize well, there generally must be a large amount of data, and the test data must be similar to the training data, allowing new answers to be interpolated in between old ones. In Krizhevsky et al's paper (Krizhevsky, Sutskever, & Hinton, 2012), a nine layer convolutional neural network with 60 million parameters and 650,000 nodes was trained on roughly a million distinct examples drawn from approximately one thousand categories.[6]

This sort of brute force approach worked well in the very finite world of ImageNet, into which all stimuli can be classified into a comparatively small set of categories. It also works well in stable domains like speech recognition in which exemplars are mapped in constant way onto a limited set of speech sound categories, but for many reasons deep learning cannot be considered (as it sometimes is in the popular press) as a general solution to artificial intelligence.

Here are ten challenges faced by current deep learning systems:

## 3.1. Deep learning thus far is data hungry

Human beings can learn abstract relationships in a few trials. If I told you that a *schmister* was a sister over the age of 10 but under the age of 21, perhaps giving you a single example, you could immediately infer whether you had any schmisters, whether your best friend had a schmister, whether your children or parents had any schmisters, and so forth. (Odds are, your parents no longer do, if they ever did, and you could rapidly draw that inference, too.)

In learning what a schmister is, in this case through explicit definition, you rely not on hundreds or thousands or millions of training examples, but on a capacity to represent abstract relationships between algebra-like variables.

Humans can learn such abstractions, both through explicit definition and more implicit means (Marcus, 2001). Indeed even 7-month old infants can do so, acquiring learned abstract language-like rules from a small number of unlabeled examples, in just two minutes (Marcus, Vijayan, Bandi Rao, & Vishton, 1999). Subsequent work by Gervain and colleagues (2012) suggests that newborns are capable of similar computations.

---

[6] Using a common technique known as data augmentation, each example was actually presented along with its label in a many different locations, both in its original form and in mirror reversed form. A second type of data augmentation varied the brightness of the images, yielding still more examples for training, in order to train the network to recognize images with different intensities. Part of the art of machine learning involves knowing what forms of data augmentation will and won't help within a given system.

Deep learning currently lacks a mechanism for learning abstractions through explicit, verbal definition, and works best when there are thousands, millions or even billions of training examples, as in DeepMind's work on board games and Atari. As Brenden Lake and his colleagues have recently emphasized in a series of papers, humans are far more efficient in learning complex rules than deep learning systems are (Lake, Salakhutdinov, & Tenenbaum, 2015; Lake, Ullman, Tenenbaum, & Gershman, 2016). (See also related work by George et al (2017), and my own work with Steven Pinker on children's overregularization errors in comparison to neural networks (Marcus et al., 1992).)

Geoff Hinton has also worried about deep learning's reliance on large numbers of labeled examples, and expressed this concern in his recent work on capsule networks with his coauthors (Sabour et al., 2017) noting that convolutional neural networks (the most common deep learning architecture) may face "exponential inefficiencies that may lead to their demise. A good candidate is the difficulty that convolutional nets have in generalizing to novel viewpoints [ie perspectives on object in visual recognition tasks]. The ability to deal with translation[al invariance] is built in, but for the other ... [common type of] transformation we have to chose between replicating feature detectors on a grid that grows exponentially ... or increasing the size of the labelled training set in a similarly exponential way."

In problems where data are limited, deep learning often is not an ideal solution.

## 3.2. Deep learning thus far is shallow and has limited capacity for transfer

Although deep learning is capable of some amazing things, it is important to realize that the word "deep" in deep learning refers to a technical, architectural property (the large number of hidden layers used in a modern neural networks, where there predecessors used only one) rather than a conceptual one (the representations acquired by such networks don't, for example, naturally apply to abstract concepts like "justice", "democracy" or "meddling").

Even more down-to-earth concepts like "ball" or "opponent" can lie out of reach. Consider for example DeepMind's Atari game work (Mnih et al., 2015) on deep reinforcement learning, which combines deep learning with reinforcement learning (in which a learner tries to maximize reward). Ostensibly, the results are fantastic: the system meets or beats human experts on a large sample of games using a single set of "hyperparameters" that govern properties such as the rate at which a network alters its weights, and no advance knowledge about specific games, or even their rules. But it is easy to wildly overinterpret what the results show. To take one example, according to a widely-circulated video of the system learning to play the brick-breaking Atari game Breakout, "after 240 minutes of training, [the system] realizes that digging a tunnel thought the wall is the most effective technique to beat the game".

But the system has learned no such thing; it doesn't really understand what a tunnel, or what a wall is; it has just learned specific contingencies for particular scenarios. *Transfer test*s — in which the deep reinforcement learning system is confronted with scenarios that differ in minor ways from the one ones on which the system was trained show that deep reinforcement learning's solutions are often extremely superficial. For example, a team of researchers at Vicarious showed that a more efficient successor technique, DeepMind's Atari system [Asynchronous Advantage Actor-Critic; also known as A3C], failed on a variety of minor perturbations to Breakout (Kansky et al., 2017) from the training set, such as moving the Y coordinate (height) of the paddle, or inserting a wall midscreen. These demonstrations make clear that it is misleading to credit deep reinforcement learning with inducing concept like wall or paddle; rather, such remarks are what comparative (animal) psychology sometimes call overattributions. It's not that the Atari system genuinely learned a concept of wall that was robust but rather the system superficially approximated breaking through walls within a narrow set of highly trained circumstances.[7]

My own team of researchers at a startup company called Geometric Intelligence (later acquired by Uber) found similar results as well, in the context of a slalom game, In 2017, a team of researchers at Berkeley and OpenAI has shown that it was not difficult to construct comparable adversarial examples in a variety of games, undermining not only DQN (the original DeepMind algorithm) but also A3C and several other related techniques (Huang, Papernot, Goodfellow, Duan, & Abbeel, 2017).

Recent experiments by Robin Jia and Percy Liang (2017) make a similar point, in a different domain: language. Various neural networks were trained on a question answering task known as SQuAD (derived from the Stanford Question Answering Database), in which the goal is to highlight the words in a particular passage that correspond to a given question. In one sample, for instance, a trained system correctly, and impressively, identified the quarterback on the winning of Super Bowl XXXIII as John Elway, based on a short paragraph. But Jia and Liang showed the mere insertion of distractor sentences (such as a fictional one about the alleged victory of Google's Jeff

---

[7] In the same paper, Vicarious proposed an alternative to deep learning called schema networks (Kansky et al., 2017) that can handle a number of variations in the Atari game Breakout, albeit apparently without the multi-game generality of DeepMind's Atari system.

Dean in another Bowl game[8]) caused performance to drop precipitously. Across sixteen models, accuracy dropped from a mean of 75% to a mean of 36%.

As is so often the case, the patterns extracted by deep learning are more superficial than they initially appear.

## 3.3. Deep learning thus far has no natural way to deal with hierarchical structure

To a linguist like Noam Chomsky, the troubles Jia and Liang documented would be unsurprising. Fundamentally, most current deep-learning based language models represent sentences as mere sequences of words, whereas Chomsky has long argued that language has a hierarchical structure, in which larger structures are recursively constructed out of smaller components. (For example, in the sentence *the teenager who previously crossed the Atlantic set a record for flying around the world*, the main clause is *the teenager set a record for flying around the world,* while the embedded clause *who previously crossed the Atlantic* is an embedded clause that specifies which teenager.)

In the 80's Fodor and Pylyshyn (1988)expressed similar concerns, with respect to an earlier breed of neural networks. Likewise, in (Marcus, 2001), I conjectured that single recurrent neural networks (SRNs; a forerunner to today's more sophisticated deep learning based recurrent neural networks, known as RNNs; Elman, 1990) would have trouble systematically representing and extending recursive structure to various kinds of unfamiliar sentences (see the cited articles for more specific claims about which types).

Earlier this year, Brenden Lake and Marco Baroni (2017) tested whether such pessimistic conjectures continued to hold true. As they put it in their title, contemporary neural nets were "Still not systematic after all these years". RNNs could "generalize well when the differences between training and test ... are small [but] when generalization requires systematic compositional skills, RNNs fail spectacularly".

Similar issues are likely to emerge in other domains, such as planning and motor control, in which complex hierarchical structure is needed, particular when a system is likely to encounter novel situations. One can see indirect evidence for this in the struggles with transfer in Atari games mentioned above, and more generally in the field of robotics, in which systems generally fail to generalize abstract plans well in novel environments.

---

[8] Here's the full Super Bowl passage; Jia and Liang's distractor sentence that confused the model is at the end. Peyton Manning became the first quarterback ever to lead two different teams to multiple Super Bowls. He is also the oldest quarterback ever to play in a Super Bowl at age 39. The past record was held by John Elway, who led the Broncos to victory in Super Bowl XXXIII at age 38 and is currently Denver's Executive Vice President of Football Operations and General Manager. Quarterback Jeff Dean had jersey number 37 in Champ Bowl XXXIV.

The core problem, at least at present, is that deep learning learns correlations between sets of features that are themselves "flat" or nonhierachical, as if in a simple, unstructured list, with every feature on equal footing. Hierarchical structure (e.g., syntactic trees that distinguish between main clauses and embedded clauses in a sentence) are not inherently or directly represented in such systems, and as a result deep learning systems are forced to use a variety of proxies that are ultimately inadequate, such as the sequential position of a word presented in a sequences.

Systems like Word2Vec (Mikolov, Chen, Corrado, & Dean, 2013) that represent individuals words as vectors have been modestly successful; a number of systems that have used clever tricks
try to represent complete sentences in deep-learning compatible vector spaces (Socher, Huval, Manning, & Ng, 2012). But, as Lake and Baroni's experiments make clear. recurrent networks continue limited in their capacity to represent and generalize rich structure in a faithful manner.

## 3.4. Deep learning thus far has struggled with open-ended inference

If you can't represent nuance like the difference between "John promised Mary to leave" and "John promised to leave Mary", you can't draw inferences about who is leaving whom, or what is likely to happen next. Current machine reading systems have achieved some degree of success in tasks like SQuAD, in which the answer to a given question is explicitly contained within a text, but far less success in tasks in which inference goes beyond what is explicit in a text, either by combining multiple sentences (so called multi-hop inference) or by combining explicit sentences with background knowledge that is not stated in a specific text selection. Humans, as they read texts, frequently derive wide-ranging inferences that are both novel and only implicitly licensed, as when they, for example, infer the intentions of a character based only on indirect dialog.

Altough Bowman and colleagues (Bowman, Angeli, Potts, & Manning, 2015; Williams, Nangia, & Bowman, 2017) have taken some important steps in this direction, there is, at present, no deep learning system that can draw open-ended inferences based on real-world knowledge with anything like human-level accuracy.

## 3.5. Deep learning thus far is not sufficiently transparent

The relative opacity of "black box" neural networks has been a major focus of discussion in the last few years (Samek, Wiegand, & Müller, 2017; Ribeiro, Singh, & Guestrin, 2016). In their current incarnation, deep learning systems have millions or even billions of parameters, identifiable to their developers not in terms of the sort of human

Page 10 of 27

interpretable labels that canonical programmers use ("last_character_typed") but only in terms of their geography within a complex network (e.g., the activity value of the $i^{th}$ node in layer $j$ in network module $k$). Although some strides have been in visualizing the contributions of individuals nodes in complex networks (Nguyen, Clune, Bengio, Dosovitskiy, & Yosinski, 2016), most observers would acknowledge that neural networks as a whole remain something of a black box.

How much that matters in the long run remains unclear (Lipton, 2016). If systems are robust and self-contained enough it might not matter; if it is important to use them in the context of larger systems, it could be crucial for debuggability.

The transparency issue, as yet unsolved, is a potential liability when using deep learning for problem domains like financial trades or medical diagnosis, in which human users might like to understand how a given system made a given decision. As Catherine O'Neill (2016) has pointed out, such opacity can also lead to serious issues of bias.

## 3.6. Deep learning thus far has not been well integrated with prior knowledge

The dominant approach in deep learning is hermeneutic, in the sense of being self-contained and isolated from other, potentially usefully knowledge. Work in deep learning typically consists of finding a training database, sets of inputs associated with respective outputs, and learn all that is required for the problem by learning the relations between those inputs and outputs, using whatever clever architectural variants one might devise, along with techniques for cleaning and augmenting the data set. With just a handful of exceptions, such as LeCun's convolutional constraint on how neural networks are wired(LeCun, 1989), prior knowledge is often deliberately minimized.

Thus, for example, in a system like Lerer et al's (2016) efforts to learn about the physics of falling towers, there is no prior knowledge of physics (beyond what is implied in convolution). Newton's laws, for example, are not explicitly encoded; the system instead (to some limited degree) approximates them by learning contingencies from raw, pixel level data. As I note in a forthcoming paper in innate (Marcus, in prep) researchers in deep learning appear to have a very strong bias against including prior knowledge even when (as in the case of physics) that prior knowledge is well known.

It also not straightforward in general how to integrate prior knowledge into a deep learning system:, in part because the knowledge represented in deep learning systems pertains mainly to (largely opaque) correlations between features, rather than to abstractions like quantified statements (e.g. *all men are mortal*), see discussion of universally-quantified one-to-one-mappings in Marcus (2001)*,* or generics (violable statements like dogs have four legs or mosquitos carry West Nile virus (Gelman, Leslie, Was, & Koch, 2015)).

A related problem stems from a culture in machine learning that emphasizes competition on problems that are inherently self-contained, without little need for broad general knowledge. This tendency is well exemplified by the machine learning contest platform known as Kaggle, in which contestants vie for the best results on a given data set. Everything they need for a given problem is neatly packaged, with all the relevant input and outputs files. Great progress has been made in this way; speech recognition and some aspects of image recognition can be largely solved in the Kaggle paradigm.

The trouble, however, is that life is not a Kaggle competition; children don't get all the data they need neatly packaged in a single directory. Real-world learning offers data much more sporadically, and problems aren't so neatly encapsulated. Deep learning works great on problems like speech recognition in which there are lots of labeled examples, but scarcely any even knows how to apply it to more open-ended problems. What's the best way to fix a bicycle that has a rope caught in its spokes? Should I major in math or neuroscience? No training set will tell us that.

Problems that have less to do with categorization and more to do with commonsense reasoning essentially lie outside the scope of what deep learning is appropriate for, and so far as I can tell, deep learning has little to offer such problems. In a recent review of commonsense reasoning, Ernie Davis and I (2015) began with a set of easily-drawn inferences that people can readily answer without anything like direct training, such as *Who is taller, Prince William or his baby son Prince George? Can you make a salad out of a polyester shirt? If you stick a pin into a carrot, does it make a hole in the carrot or in the pin?*

As far as I know, nobody has even tried to tackle this sort of thing with deep learning.

Such apparently simple problems require humans to integrate knowledge across vastly disparate sources, and as such are a long way from the sweet spot of deep learning-style perceptual classification. Instead, they are perhaps best thought of as a sign that entirely different sorts of tools are needed, along with deep learning, if we are to reach human-level cognitive flexibility.

### 3.7. Deep learning thus far cannot inherently distinguish causation from correlation

If it is a truism that causation does not equal correlation, the distinction between the two is also a serious concern for deep learning. Roughly speaking, deep learning learns complex correlations between input and output features, but with no inherent representation of causality. A deep learning system can easily learn that height and vocabulary are, across the population as a whole, correlated, but less easily represent the way in which that correlation derives from growth and development (kids get bigger as they learn more words, but that doesn't mean that growing tall causes them to learn more words, nor that learning new words causes them to grow). Causality has been central strand in some other approaches to AI (Pearl, 2000) but, perhaps because deep learning is not geared towards such challenges, relatively little work within the deep learning tradition has tried to address it.[9]

## 3.8. Deep learning presumes a largely stable world, in ways that may be problematic

The logic of deep learning is such that it is likely to work best in highly stable worlds, like the board game Go, which has unvarying rules, and less well in systems such as politics and economics that are constantly changing. To the extent that deep learning is applied in tasks such as stock prediction, there is a good chance that it will eventually face the fate of Google Flu Trends, which initially did a great job of predicting epidemiological data on search trends, only to complete miss things like the peak of the 2013 flu season (Lazer, Kennedy, King, & Vespignani, 2014).

## 3.9. Deep learning thus far works well as an approximation, but its answers often cannot be fully trusted

In part as a consequence of the other issues raised in this section, deep learning systems are quite good at some large fraction of a given domain, yet easily fooled.

An ever-growing array of papers has shown this vulnerability, from the linguistic examples of Jia and Liang mentioned above to a wide range of demonstrations in the domain of vision, where deep learning systems have mistaken yellow-and-black patterns of stripes for school buses (Nguyen, Yosinski, & Clune, 2014) and sticker-clad parking signs for well-stocked refrigerators (Vinyals, Toshev, Bengio, & Erhan, 2014) in the context of a captioning system that otherwise seems impressive.

More recently, there have been real-world stop signs, lightly defaced, that have been mistaken for speed limit signs (Evtimov et al., 2017) and 3d-printed turtles that have been mistake for rifles (Athalye, Engstrom, Ilyas, & Kwok, 2017). A recent news story

---

[9] One example of interesting recent work is (Lopez-Paz, Nishihara, Chintala, Schölkopf, & Bottou, 2017), albeit focused specifically on an rather unusual sense of the term causation as it relates to the presence or absence of objects (e.g., "the presence of cars cause the presence of wheel[s]). This strikes me as quite different from the sort of causation one finds in the relation between a disease and the symptoms it causes.

recounts the trouble a British police system has had in distinguishing nudes from sand dunes.[10]

The "spoofability" of deep learning systems was perhaps first noted by Szegedy et al(2013). Four years later, despite much active research, no robust solution has been found.[11]

## 3.10. Deep learning thus far is difficult to engineer with

Another fact that follows from all the issues raised above is that is simply hard to do robust engineering with deep learning. As a team of authors at Google put it in 2014, in the title of an important, and as yet unanswered essay (Sculley, Phillips, Ebner, Chaudhary, & Young, 2014), machine learning is "the high-interest credit card of technical debt", meaning that is comparatively easy to make systems that work in some limited set of circumstances (short term gain), but quite difficult to guarantee that they will work in alternative circumstances with novel data that may not resemble previous training data (long term debt, particularly if one system is used as an element in another larger system).

In an important talk at ICML, Leon Bottou (2015) compared machine learning to the development of an airplane engine, and noted that while the airplane design relies on building complex systems out of simpler systems for which it was possible to create sound guarantees about performance, machine learning lacks the capacity to produce comparable guarantees. As Google's Peter Norvig (2016) has noted, machine learning as yet lacks the incrementality, transparency and debuggability of classical programming, trading off a kind of simplicity for deep challenges in achieving robustness.

Henderson and colleagues have recently extended these points, with a focus on deep reinforcement learning, noting some serious issues in the field related to robustness and replicability (Henderson et al., 2017).

Although there has been some progress in automating the process of developing machine learning systems (Zoph, Vasudevan, Shlens, & Le, 2017), there is a long way to go.

---

[10] https://gizmodo.com/british-cops-want-to-use-ai-to-spot-porn-but-it-keeps-m-1821384511/amp

[11] Deep learning's predecessors were vulnerable to similar problems, as Pinker and Prince (1988)pointed out, in a discussion of neural networks that produced bizarre past tense forms for a subset of its inputs. The verb to *mail*, for example, was inflected in the past tense as *membled*, the verb *tour* as *toureder*. Children rarely if ever make mistakes like these.

## 3.11. Discussion

Of course, deep learning, is by itself, just mathematics; none of the problems identified above are because the underlying mathematics of deep learning are somehow flawed. In general, deep learning is a perfectly fine way of optimizing a complex system for representing a mapping between inputs and outputs, given a sufficiently large data set.

The real problem lies in misunderstanding what deep learning is, and is not, good for. The technique excels at solving closed-end classification problems, in which a wide range of potential signals must be mapped onto a limited number of categories, given that there is enough data available and the test set closely resembles the training set.

But deviations from these assumptions can cause problems; deep learning is just a statistical technique, and all statistical techniques suffer from deviation from their assumptions.

Deep learning systems work less well when there are limited amounts of training data available, or when the test set differs importantly from the training set, or when the space of examples is broad and filled with novelty. And some problems cannot, given real-world limitations, be thought of as classification problems at all. Open-ended natural language understanding, for example, should not be thought of as a classifier mapping between a large finite set of sentences and large, finite set of sentences, but rather a mapping between a potentially infinite range of input sentences and an equally vast array of meanings, many never previously encountered. In a problem like that, deep learning becomes a square peg slammed into a round hole, a crude approximation when there must be a solution elsewhere.

One clear way to get an intuitive sense of why something is amiss to consider a set of experiments I did long ago, in 1997, when I tested some simplified aspects of language development on a class of neural networks that were then popular in cognitive science. The 1997-vintage networks were, to be sure, simpler than current models — they used no more than three layers (inputs nodes connected to hidden nodes connected to outputs node), and lacked Lecun's powerful convolution technique. But they were driven by backpropagation just as today's systems are, and just as beholden to their training data.

In language, the name of the game is generalization — once I hear a sentence like *John pilked a football to Mary*, I can infer that is also grammatical to say *John pilked Mary the football*, and *Eliza pilked the ball to Alec;* equally if I can infer what the word *pilk* means, I can infer what the latter sentences would mean, even if I had not hear them before.

Distilling the broad-ranging problems of language down to a simple example that I believe still has resonance now, I ran a series of experiments in which I trained three-layer perceptrons (fully connected in today's technical parlance, with no convolution) on the identity function, *f(x) = x,* e.g, *f(12)=12*.

Training examples were represented by a set of input nodes (and corresponding output nodes) that represented numbers in terms of binary digits. The number 7 for example, would be represented by turning on the input (and output) nodes representing 4, 2, and 1. As a test of generalization, I trained the network on various sets of even numbers, and tested it all possible inputs, both odd and even.

Every time I ran the experiment, using a wide variety of parameters, the results were the same: the network would (unless it got stuck in local minimum) correctly apply the identity function to the even numbers that it had seen before (say 2, 4, 8 and 12), and to some other even numbers (say 6 and 14) but fail on all the odds numbers, yielding, for example f(15) = 14.

In general, the neural nets I tested could learn their training examples, and interpolate to a set of test examples that were in a cloud of points around those examples in n-dimensional space (which I dubbed the *training space*), but they could not extrapolate beyond that training space.

Odd numbers were outside the training space, and the networks could not generalize identity outside that space.[12] Adding more hidden units didn't help, and nor did adding more hidden layers. Simple multilayer perceptrons simply couldn't generalize outside their training space (Marcus, 1998a; Marcus, 1998b; Marcus, 2001). (Chollet makes quite similar points in the closing chapters of his his (Chollet, 2017) text.)

What we have seen in this paper is that challenges in generalizing beyond a space of training examples persist in current deep learning networks, nearly two decades later. Many of the problems reviewed in this paper — the data hungriness, the vulnerability to fooling, the problems in dealing with open-ended inference and transfer —   can be seen as extension of this fundamental problem. Contemporary neural networks do well on challenges that remain close to their core training data, but start to break down on cases further out in the periphery.

---

[12] Of course, the network had never seen an odd number before, but pretraining the network on odd numbers in a different context didn't help. And of course people, in contrast, readily generalize to novel words immediately upon hearing them. Likewise, the experiments I did with seven-month-olds consisted entirely of novel words.

The widely-adopted addition of convolution guarantees that one particular class of problems that are akin to my identity problem can be solved: so-called translational invariances, in which an object retains its identity when it is shifted to a location. But the solution is not general, as for example Lake's recent demonstrations show. (Data augmentation offers another way of dealing with deep learning's challenges in extrapolation, by trying to broaden the space of training examples itself, but such techniques are more useful in 2d vision than in language).

As yet there is no general solution within deep learning to the problem of generalizing outside the training space. And it is for that reason, more than any other, that we need to look to different kinds of solutions if we want to reach artificial general intelligence.

## 4. Potential risks of excessive hype

One of the biggest risks in the current overhyping of AI is another AI winter, such as the one that devastated the field in the 1970's, after the Lighthill report (Lighthill, 1973), suggested that AI was too brittle, too narrow and too superficial to be used in practice. Although there are vastly more practical applications of AI now than there were in the 1970s, hype is still a major concern. When a high-profile figure like Andrew Ng writes in the Harvard Business Review promising a degree of imminent automation that is out of step with reality, there is fresh risk for seriously dashed expectations. Machines cannot in fact do many things that ordinary humans can do in a second, ranging from reliably comprehending the world to understanding sentences. No healthy human being would ever mistake a turtle for a rifle or parking sign for a refrigerator.

Executives investing massively in AI may turn out to be disappointed, especially given the poor state of the art in natural language understanding. Already, some major projects have been largely abandoned, like Facebook's M project, which was launched in August 2015 with much publicity[13] as a general purpose personal assistant, and then later downgraded to a significantly smaller role, helping users with a vastly small range of well-defined tasks such as calendar entry.

It is probably fair to say that chatbots in general have not lived up to the hype they received a couple years ago. If, for example, driverless car should also, disappoint, relative to their early hype, by proving unsafe when rolled out at scale, or simply not achieving full autonomy after many promises, the whole field of AI could be in for a sharp downturn, both in popularity and funding. We already may be seeing hints of this, as in a just published Wired article[14] that was entitled "After peak hype, self-driving cars enter the trough of disillusionment."

---

[13] https://www.wired.com/2015/08/how-facebook-m-works/

There are other serious fears, too, and not just of the apocalyptic variety (which for now to still seem to be stuff of science fiction). My own largest fear is that the field of AI could get trapped in a local minimum, dwelling too heavily in the wrong part of intellectual space, focusing too much on the detailed exploration of a particular class of accessible but limited models that are geared around capturing low-hanging fruit — potentially neglecting riskier excursions that might ultimately lead to a more robust path.

I am reminded of Peter Thiel's famous (if now slightly outdated) damning of an often too-narrowly focused tech industry: "We wanted flying cars, instead we got 140 characters". I still dream of Rosie the Robost, a full-service domestic robot that take of my home; but for now, six decades into the history of AI, our bots do little more than play music, sweep floors, and bid on advertisements.

If didn't make more progress, it would be a shame. AI comes with risk, but also great potential rewards. AI's greatest contributions to society, I believe, could and should ultimately come in domains like automated scientific discovery, leading among other things towards vastly more sophisticated versions of medicine than are currently possible. But to get there we need to make sure that the field as whole doesn't first get stuck in a local minimum.

## 5. What would be better?

Despite all of the problems I have sketched, I don't think that we need to abandon deep learning.

Rather, we need to reconceptualize it: not as a universal solvent, but simply as one tool among many, a power screwdriver in a world in which we also need hammers, wrenches, and pliers, not to mentions chisels and drills, voltmeters, logic probes, and oscilloscopes.

In perceptual classification, where vast amounts of data are available, deep learning is a valuable tool; in other, richer cognitive domains, it is often far less satisfactory.

The question is, where else should we look? Here are four possibilities.

---

[14] https://www.wired.com/story/self-driving-cars-challenges/

## 5.1. Unsupervised learning

In interviews, deep learning pioneers Geoff Hinton and Yann LeCun have both recently pointed to unsupervised learning as one key way in which to go beyond supervised, data-hungry versions of deep learning.

To be clear, deep learning and unsupervised learning are not in logical opposition. Deep learning has mostly been used in a supervised context with labeled data, but there are ways of using deep learning in an unsupervised fashion. But there is certainly reasons in many domains to move away from the massive demands on data that supervised deep learning typically requires.

Unsupervised learning, as the term is commonly used, tends to refer to several kinds of systems. One common type of system "clusters" together inputs that share properties, even without having them explicitly labeled. Google's cat detector model (Le et al., 2012) is perhaps the most publicly prominent example of this sort of approach.

Another approach, advocated researchers such as Yann LeCun (Luc, Neverova, Couprie, Verbeek, & LeCun, 2017), and not mutually exclusive with the first, is to replace labeled data sets with things like movies that change over time. The intuition is that systems trained on videos can use each pair of successive frames as a kind of ersatz teaching signal, in which the goal is to predict the next frame; frame $t$ becomes a predictor for frame $t1$, without the need for any human labeling.

My view is that both of these approaches are useful (and so are some others not discussed here), but that neither inherently solve the sorts of problems outlined in section 3. One is still left with data hungry systems that lack explicit variables, and I see no advance there towards open-ended inference, interpretability or debuggability.

That said, there is a different notion of unsupervised learning, less discussed, which I find deeply interesting: the kind of unsupervised learning that human children do. Children often y set themselves a novel task, like creating a tower of Lego bricks or climbing through a small aperture, as my daughter recently did in climbing through a chair, in the space between the seat and the chair back . Often, this sort of exploratory problem solving involves (or at least appears to involve) a good deal of autonomous goal setting (what should I do?) and high level problem solving (how do I get my arm through the chair, now that the rest of my body has passed through?), as well the integration of abstract knowledge (how bodies work, what sorts of apertures and affordances various objects have, and so forth). If we could build systems that could set their own goals and do reasoning and problem-solving at this more abstract level, major progress might quickly follow.

## 5.2. Symbol-manipulation, and the need for hybrid models

Another place that we should look is towards classic, "symbolic" AI, sometimes referred to as GOFAI (Good Old-Fashioned AI). Symbolic AI takes its name from the idea, central to mathematics, logic, and computer science, that abstractions can be represented by symbols. Equations like f = ma allow us to calculate outputs for a wide range of inputs, irrespective of whether we have seen any particular values before; lines in computer programs do the same thing (if the value of variable *x* is greater than the value of variable *y,* perform action *a*).

By themselves, symbolic systems have often proven to be brittle, but they were largely developed in era with vastly less data and computational power than we have now. The right move today may be to integrate deep learning, which excels at perceptual classification, with symbolic systems, which excel at inference and abstraction. One might think such a potential merger on analogy to the brain; perceptual input systems, like primary sensory cortex, seem to do something like what deep learning does, but there are other areas, like Broca's area and prefrontal cortex, that seem to operate at much higher level of abstraction. The power and flexibility of the brain comes in part from its capacity to dynamically *integrate* many different computations in real-time. The process of scene perception, for instance, seamlessly integrates direct sensory information with complex abstractions about objects and their properties, lighting sources, and so forth.

Some tentative steps towards integration already exist, including neurosymbolic modeling (Besold et al., 2017) and recent trend towards systems such as differentiable neural computers (Graves et al., 2016), programming with differentiable interpreters (Bošnjak, Rocktäschel, Naradowsky, & Riedel, 2016), and neural programming with discrete operations (Neelakantan, Le, Abadi, McCallum, & Amodei, 2016). While none of this work has yet fully scaled towards anything like full-service artificial general intelligence, I have long argued (Marcus, 2001) that more on integrating microprocessor-like operations into neural networks could be extremely valuable.

To the extent that the brain might be seen as consisting of "a broad array of reusable computational primitives—elementary units of processing akin to sets of basic instructions in a microprocessor—perhaps wired together in parallel, as in the reconfigurable integrated circuit type known as the field-programmable gate array", as I have argued elsewhere(Marcus, Marblestone, & Dean, 2014), steps towards enriching the instruction set out of which our computational systems are built can only be a good thing.

# 5.3. More insight from cognitive and developmental psychology

Another potential valuable place to look is human cognition (Davis & Marcus, 2015; Lake et al., 2016; Marcus, 2001; Pinker & Prince, 1988). There is no need for machines to literally replicate the human mind, which is, after all, deeply error prone, and far from perfect. But there remain many areas, from natural language understanding to commonsense reasoning, in which humans still retain a clear advantage; learning the mechanisms underlying those human strengths could lead to advances in AI, even the goal is not, and should not be, an exact replica of human brain.

For many people, learning from humans means neuroscience; in my view, that may be premature. We don't yet know enough about neuroscience to literally reverse engineer the brain, per se, and may not for several decades, possibly until AI itself gets better. AI can help us to decipher the brain, rather than the other way around.

Either way, in the meantime, it should certainly be possible to use techniques and insights drawn from cognitive and developmental and psychology, now, in order to build more robust and comprehensive artificial intelligence, building models that are motivated not just by mathematics but also by clues from the strengths of human psychology.

A good starting point might be to first to try understand the innate machinery in humans minds, as a source of hypotheses into mechanisms that might be valuable in developing artificial intelligences; in companion article to this one (Marcus, in prep) I summarize a number of possibilities, some drawn from my own earlier work (Marcus, 2001) and others from Elizabeth Spelke's (Spelke & Kinzler, 2007). Those drawn from my own work focus on how information might be represented and manipulated, such as by symbolic mechanisms for representing variables and distinctions between kinds and individuals from a class; those drawn from Spelke focus on how infants might represent notions such as space, time, and object.

A second focal point might be on common sense knowledge, both in how it develops (some might be part of our innate endowment, much of it is learned), how it is represented, and how it is integrated on line in the process of our interactions with the real world (Davis & Marcus, 2015). Recent work by Lerer et al (2016), Watters and colleagues (2017), Tenenbaum and colleagues (Wu, Lu, Kohli, Freeman, & Tenenbaum, 2017) and Davis and myself (Davis, Marcus, & Frazier-Logue, 2017) suggest some competing approaches to how to think about this, within the domain of everyday physical reasoning.

A third focus might be on human understanding of narrative, a notion long ago suggested by Roger Schank and Abelson (1977) and due for a refresh (Marcus, 2014; Kočiský et al., 2017).

### 5.4. Bolder challenges

Whether deep learning persists in current form, morphs into something new, or gets replaced altogether, one might consider a variety of challenge problems that push systems to move beyond what can be learned in supervised learning paradigms with large datasets. Drawing in part of from a recent special issue of AI Magazine devoted to moving beyond the Turing Test that I edited with Francesca Rossi, Manuelo Veloso (Marcus, Rossi, Veloso - AI Magazine, & 2016, 2016), here are a few suggestions:

- A comprehension challenge (Paritosh & Marcus, 2016; Kočiský et al., 2017)] which would require a system to watch an arbitrary video (or read a text, or listen to a podcast) and answer open-ended questions about what is contained therein. (Who is the protagonist? What is their motivation? What will happen if the antagonist succeeds in her mission?) No specific supervised training set can cover all the possible contingencies; infererence and real-world knowledge integration are necessities.
- Scientific reasoning and understanding, as in the Allen AI institute's 8th grade science challenge (Schoenick, Clark, Tafjord, P, & Etzioni, 2017; Davis, 2016). While the answers to many basic science questions can simply be retrieved from web searches, others require inference beyond what is explicitly stated, and the integration of general knowledge.
- General game playing (Genesereth, Love, & Pell, 2005), with transfer between games (Kansky et al., 2017), such that, for example, learning about one first-person shooter enhances performance on another with entirely different images, equipment and so forth. (A system that can learn many games, separately, without transfer between them, such as DeepMind's Atari game system, would not qualify; the point is to acquire cumulative, transferrable knowledge).
- A physically embodied test an AI-driven robot that could build things (Ortiz Jr, 2016), ranging from tents to IKEA shelves, based on instructions and real-world physical interactions with the objects parts, rather than vast amounts trial-and-error.

No one challenge is likely to be sufficient. Natural intelligence is multi-dimensional (Gardner, 2011), and given the complexity of the world, generalized artificial intelligence will necessarily be multi-dimensional as well.

By pushing beyond perceptual classification and into a broader integration of inference and knowledge, artificial intelligence will advance, greatly.

# 6. Conclusions

As a measure of progress, it is worth considering a somewhat pessimistic piece I wrote for *The New Yorker* five years ago[15], conjecturing that "deep learning is only part of the larger challenge of building intelligent machines" because "such techniques lack ways of representing causal relationships (such as between diseases and their symptoms), and are likely to face challenges in acquiring abstract ideas like "sibling" or "identical to." They have no obvious ways of performing logical inferences, and they are also still a long way from integrating abstract knowledge, such as information about what objects are, what they are for, and how they are typically used."

As we have seen, many of these concerns remain valid, despite major advances in specific domains like speech recognition, machine translation, and board games, and despite equally impressive advances in infrastructure and the amount of data and compute available.

Intriguingly, in the last year, a growing array of other scholars, coming from an impressive range of perspectives, have begun to emphasize similar limits. A partial list includes Brenden Lake and Marco Baroni (2017), François Chollet (2017), Robin Jia and Percy Liang (2017), Dileep George and others at Vicarious (Kansky et al., 2017) and Pieter Abbeel and colleagues at Berkeley (Stoica et al., 2017).

Perhaps most notably of all, Geoff Hinton has been courageous enough to reconsider has own beliefs, revealing in an August interview with the news site Axios[16] that he is "deeply suspicious" of back-propagation, a key enabler of deep learning that he helped pioneer, because of his concern about its dependence on labeled data sets.

Instead, he suggested (in Axios' paraphrase) that "entirely new methods will probably have to be invented."

I share Hinton's excitement in seeing what comes next.

---

[15] https://www.newyorker.com/news/news-desk/is-deep-learning-a-revolution-in-artificial-intelligence

[16] https://www.axios.com/ai-pioneer-advocates-starting-over-2485537027.html

Page 23 of 27